\documentclass{article} 
\usepackage[preprint]{colm2026_conference}
\usepackage[T1]{fontenc}
\usepackage{microtype}
\usepackage{graphicx}
\usepackage{subcaption}
\usepackage{booktabs}
\usepackage{multirow}
\usepackage{hyperref}
\usepackage{url}
\usepackage{enumitem}
\usepackage{lineno}

\usepackage{algorithm}
\usepackage{algorithmic}

\definecolor{darkblue}{rgb}{0, 0, 0.5}
\hypersetup{colorlinks=true, citecolor=darkblue, linkcolor=darkblue, urlcolor=darkblue}

\usepackage{amsmath}
\usepackage{amssymb}
\usepackage{mathtools}
\usepackage{amsthm}
\newcommand{\ind}{\mathbf{1}}
\usepackage[capitalize,noabbrev]{cleveref}

\usepackage[textsize=tiny,backgroundcolor=yellow!30]{todonotes}
\presetkeys{todonotes}{inline}{}  

\usepackage{xcolor}
\definecolor{draftblue}{RGB}{220, 240, 255}

\usepackage{pifont}

\usepackage{wrapfig}
\usepackage{tcolorbox}
\usepackage{listings}

\lstdefinelanguage{json}{
    basicstyle=\ttfamily\scriptsize,
    numbers=none,
    numberstyle=\scriptsize,
    stepnumber=1,
    numbersep=8pt,
    showstringspaces=false,
    breaklines=true,
    breakatwhitespace=true,
    breakindent=0pt,
    frame=none,
    backgroundcolor=\color{gray!5},
    literate=
     *{0}{{{\color{blue}0}}}{1}
      {1}{{{\color{blue}1}}}{1}
      {2}{{{\color{blue}2}}}{1}
      {3}{{{\color{blue}3}}}{1}
      {4}{{{\color{blue}4}}}{1}
      {5}{{{\color{blue}5}}}{1}
      {6}{{{\color{blue}6}}}{1}
      {7}{{{\color{blue}7}}}{1}
      {8}{{{\color{blue}8}}}{1}
      {9}{{{\color{blue}9}}}{1}
      {:}{{{\color{red}{:}}}}{1}
      {,}{{{\color{red}{,}}}}{1}
      {\{}{{{\color{black}{\{}}}}{1}
      {\}}{{{\color{black}{\}}}}}{1}
      {[}{{{\color{black}{[}}}}{1}
      {]}{{{\color{black}{]}}}}{1},
}

\definecolor{Turquoise}{RGB}{15, 76, 129}
\NewDocumentCommand{\haoyang}{ mO{} }{
\textcolor{Turquoise}{\textsuperscript{\textsc{Haoyang}}\textsf{\textbf{\small[#1]}}}}

\title{Training LLMs for Multi-Step Tool Orchestration with \\
  Constrained Data Synthesis and Graduated Rewards}

\author{Jiayang Cheng$^{1,2}$\thanks{$^{1}$Amazon Services Inc., Palo Alto, CA, USA. $^{2}$Department of Computer Science and Engineering, HKUST, Hong Kong SAR, China. $^{3}$School of Computational Science and Engineering, Georgia Institute of Technology, Atlanta, GA, USA. Correspondence to: Jiayang Cheng (during his internship at Amazon)<jchengaj@cse.ust.hk>, Xin Liu <xliucr@amazon.com>.} \And Xin Liu$^{1}$ \And Zhihan Zhang$^{1}$ \And Haoyang Wen$^{1}$ \AND Zixuan Zhang$^{1}$ \And Qingyu Yin$^{1}$ \And Shiyang Li$^{1}$ \And Priyanka Nigam$^{1}$ \AND Bing Yin$^{1}$ \And Chao Zhang$^{1,3}$ \And Yangqiu Song$^{1,2}$ 
}

\begin{document}

\ifcolmsubmission
\linenumbers
\fi

\maketitle

\begin{abstract}
Multi-step tool orchestration remains challenging for LLMs, as state-of-the-art models frequently fail on full sequence execution due to parameter errors. Training for these workflows faces two obstacles: the lack of environments supporting complex real-world API dependencies, and sparse binary rewards that provide no signal for partial correctness. We propose a reinforcement learning framework addressing both challenges. First, we construct a deterministic environment backed by a large-scale cache of real API responses, enabling constrained synthesis of valid multi-step traces with controllable complexity. Second, we introduce a graduated reward that decomposes correctness into atomic validity (call-level correctness at increasing granularity) and orchestration consistency (correct sequencing with dependency respect). On ComplexFuncBench, our approach substantially improves turn accuracy, with ablations confirming both reward components are essential. Cross-benchmark evaluation on BFCL v4 shows that the learned orchestration skills transfer to entirely different API ecosystems (e.g., agentic web search and memory management), yielding consistent gains while maintaining stable single-step performance. Code is available at \url{https://github.com/horizon-rl/ToolOrchestrationReward}.
\end{abstract}

\section{Introduction}
\label{sec:intro}

Large language models (LLMs) have demonstrated remarkable capabilities in using external tools through function calling~\citep{gorilla,toolllm}. While significant progress has been made on single-step tool invocation, real-world applications often require multi-step tool orchestration, i.e., invoking a sequence of APIs in the correct order while propagating outputs from earlier calls as inputs to subsequent ones.

Recent benchmarks reveal that even state-of-the-art models struggle with multi-step orchestration. On ComplexFuncBench~\citep{complexfuncbench}, parameter value errors dominate failure modes (up to 78.8\% for Qwen2.5-72B), indicating that models often select the correct function but fail to infer precise parameter values. Similarly, NESTFUL~\citep{nestful} reports that the best models achieve only 28\% full sequence accuracy on nested function calls, with performance degrading sharply at depth $\geq 2$.

These findings point to two fundamental obstacles for applying reinforcement learning (RL) to multi-step tool orchestration. \textbf{Challenge 1} concerns the lack of suitable training environments. While execution environments exist for tool learning (e.g., BFCL~\citep{bfcl}, $\tau$-bench~\citep{yao2024tau}), they primarily target multi-user-turn scenarios with simple per-turn function calls rather than complex multi-step orchestration where a single user query requires multiple dependent API calls. Moreover, these environments often rely on simulated or synthetic data rather than real API responses, which can produce inconsistent dependency chains. \textbf{Challenge 2} concerns sparse rewards. Existing RL-based tool-use frameworks, such as ToolRL~\citep{toolrl} and PARL-MT~\citep{parlmt}, primarily rely on terminal success signals or binary execution feedback, providing no signal for partially correct reasoning paths. RAGEN~\citep{ragen} identifies this issue but focuses on reasoning tasks rather than multi-step tool use.

\begin{figure*}[t]
\centering
\includegraphics[width=0.96\linewidth]{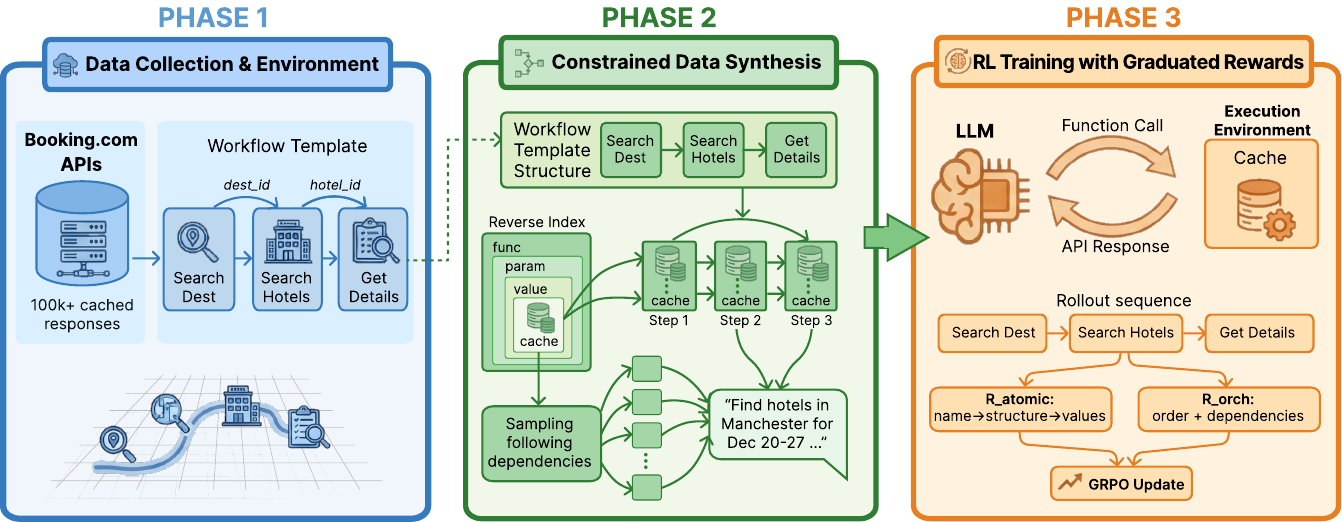}
\caption{Framework overview using hotel booking as a running example. \textbf{Phase 1}: Collect real API responses and curate workflow templates with dependency structures. \textbf{Phase 2}: Sample cache entries following dependencies, then generate queries matching sampled parameters. \textbf{Phase 3}: LLM interacts with the deterministic environment through multi-turn rollouts, receiving graduated rewards ($R_{\text{atomic}}$ + $R_{\text{orch}}$) for GRPO updates.}
\label{fig:overview}
\end{figure*}

We present a systematic framework addressing both challenges (\Cref{fig:overview}):

\begin{enumerate}[leftmargin=*]
    \item \textbf{Deterministic RL Training Environment with Constrained Data Synthesis.} We construct a cache-based execution environment with 100k+ real API responses, ensuring consistent dependency chains across training iterations. Our constrained synthesis pipeline leverages workflow templates to achieve high generation success rates.

    \item \textbf{Graduated Reward Design.} We decompose correctness into $R_{\text{atomic}}$ (validating individual function calls at increasing granularity) and $R_{\text{orch}}$ (validating correct sequencing with dependency respect), providing dense learning signals for complex trajectories.

    \item \textbf{Empirical Validation.} We evaluate in two settings: training on synthetic data to test generalization, and an oracle setting (training on in-domain evaluation queries) to establish an upper bound. RL training demonstrates substantial improvements in turn accuracy, with ablation studies confirming both reward components are essential. Cross-benchmark evaluation on BFCL v4 shows that the learned orchestration capabilities transfer to entirely different API ecosystems, improving agentic web search accuracy from 3.5\% to 10.5\% and memory tasks from 17.4\% to 21.7\%.
\end{enumerate}

\section{Methodology}
\label{sec:method}

\subsection{Problem Formulation}
\label{sec:formulation}

We formulate multi-step tool orchestration as a sequential decision-making problem. Given a user query $q$ and a set of available tools $\mathcal{F}$, the agent must generate a sequence of actions $\mathbf{y} = (y_1, \dots, y_T)$, where each action $y_t = (f_t, \boldsymbol{\theta}_t)$ consists of a function selection $f_t \in \mathcal{F}$ and parameter assignment $\boldsymbol{\theta}_t$.

The execution environment is deterministic. Executing $y_t$ yields an observation $o_t = \text{Exec}(f_t, \boldsymbol{\theta}_t)$. Crucially, a subsequent action $y_{t'}$ (where $t' > t$) may possess a \textit{dependency} on previous outputs. This is modeled as $\boldsymbol{\theta}_{t'} \leftarrow \phi(o_{1:t'-1})$, where $\phi$ represents the \textit{ground truth} logic for extracting and propagating values (e.g., extracting a \texttt{hotel\_id} to query room details). The agent must learn to approximate this dependency logic.

Our objective is to learn a policy $\pi_\theta(\mathbf{y}|q)$ that maximizes a reward $R(\mathbf{y}, \mathbf{y}^*)$, where $\mathbf{y}^*$ represents the ground truth workflow. Unlike standard generation tasks, validity is constrained by the underlying API logic: an action $y_t$ is valid if and only if all its dependencies are satisfied by the history $H_{t-1} = \{(y_k, o_k)\}_{k=1}^{t-1}$.

\subsection{Deterministic Training Environment}
\label{sec:environment}
Training LLMs for multi-step tool orchestration via RL requires a suitable execution environment. Existing approaches like BFCL~\citep{bfcl} or $\tau$-bench~\citep{yao2024tau} often rely on simulated responses or lack complex intra-turn dependencies. We address this by constructing a deterministic environment grounded in real data. Our construction process follows three steps: defining workflow logic, collecting targeted data, and enabling constrained synthesis.

\begin{itemize}[leftmargin=*]
    \item \textbf{Workflow Template Curation.}
    Before collecting data, we must define valid orchestration patterns. We formalize recurring patterns as \textit{Workflow Templates} that capture: ordered function sequences $(f_1, \dots, f_T)$, dependency structures specifying how $\boldsymbol{\theta}_t$ depends on previous observations $o_{<t}$, and field paths for value extraction (implementing $\phi$). Templates encode the inherent dependency structure of the API ecosystem, formalizing which functions must precede others and how outputs flow as inputs, analogous to API documentation. For a new domain, templates can be systematically derived from API specifications by analyzing input/output parameter dependencies between endpoints. We curate over 100 templates spanning five domains (see \Cref{app:workflow-template} in the appendix for a concrete example).

    \item \textbf{Workflow-Aware Cache Collection.}
    To enable deterministic RL training without the latency and instability of live API calls, we build a cache-based environment that implements $o_t = \text{Exec}(f_t, \boldsymbol{\theta}_t)$. Crucially, we employ \textit{workflow-aware collection}: rather than caching random independent API calls, we execute complete action sequences $(y_1, \dots, y_T)$ from the templates against live Booking.com APIs. This ensures that for every cached observation $o_t$, the corresponding dependent observations $o_{t+1}, \dots$ that rely on values extracted from $o_t$ are also present.

    We collected 100k+ real API responses covering 40 functions in $\mathcal{F}$. Responses are indexed by hash of $(f, \boldsymbol{\theta})$, enabling $O(1)$ deterministic lookups. We implement comprehensive input validation with 8 specialized validators (date formats, coordinate ranges, enum values) to return structured error messages matching real API behavior. During RL rollouts, if the model generates a call not present in the cache, the environment returns a structured error response (e.g., \texttt{\{"error": "No results found"\}}), mirroring real API behavior for invalid queries without leaking implementation details.

    \item \textbf{Constrained Data Synthesis.} Rather than caching function responses independently, we execute complete workflow chains (\textit{workflow-aware cache expansion}) to ensure cached responses form valid multi-step traces.
    The key insight is that freely exploring function chains leads to low data generation efficiency, as arbitrary combinations often violate dependency constraints or produce unsatisfiable parameter bindings. Instead, we use workflow templates (pre-validated on live APIs) as skeletons, then sample cache entries step-by-step following the dependency structure, and finally prompt an LLM to synthesize a query $q$ that would elicit the sampled action sequence $\{(f_t, \boldsymbol{\theta}_t)\}_{t=1}^T$. The pipeline works as follows:

   \textbf{(1) Reverse Index Construction.}
   We build a three-level inverted index $\mathcal{I}: (f, \text{param}, \text{val}) \to \{\text{cache\_ids}\}$. For example, $\mathcal{I}(\texttt{Search\_Hotels}, \texttt{dest\_id}, \texttt{"ABC123"})$ returns all cache entries where \texttt{Search\_Hotels} was called with \texttt{dest\_id="ABC123"}.

    \textbf{(2) Constraint-Aware Cache Sampling.}
    For each workflow template, we sample action-observation pairs $(y_t, o_t)$ step-by-step from the cache. Independent steps (no dependencies) sample $\boldsymbol{\theta}_t$ randomly from cached entries for $f_t$. Dependent steps use the reverse index: extract the required value from $o_{t-1}$, query the index, and intersect with other constraints to obtain valid $\boldsymbol{\theta}_t$. If the intersection is empty (no compatible cache entries), sampling restarts from step 1 with different random choices. After a fixed number of retries, the current attempt is abandoned and a new workflow is sampled. This guarantees all successfully sampled entries form a valid ground truth sequence $\mathbf{y}^*$.

    \textbf{(3) Query Generation.}
    Given the sampled ground truth $\mathbf{y}^* = \{(f_t, \boldsymbol{\theta}_t)\}_{t=1}^T$, we prompt an LLM to generate a query $q$ that would elicit this action sequence. The prompt design prioritizes \textit{parameter fidelity} over linguistic diversity: we instruct the model to incorporate all $\boldsymbol{\theta}_t$ values verbatim and require it to echo back the parameters in structured JSON format alongside $q$. This echo-back mechanism enables automatic consistency verification between the natural language surface form and the underlying parameter bindings. We employ zero-shot prompting without few-shot exemplars, relying on the instruction-following capabilities of modern LLMs. The complete prompt template is provided in \Cref{app:prompt-templates} in the appendix.

    \textbf{(4) Validation.}
    We execute each $y_t^* = (f_t, \boldsymbol{\theta}_t)$ through the cache-based environment and verify: (a) cache hit for $(f_t, \boldsymbol{\theta}_t)$, (b) successful execution yielding valid $o_t$. Since parameters are sampled directly from cache, validation serves as a sanity check. Any step failure invalidates the entire $(q, \mathbf{y}^*)$ pair.
\end{itemize}

The complete synthesis process is formalized in Algorithm~\ref{alg:synthesis} (\Cref{app:algorithm}), with a concrete training sample provided in \Cref{app:training-sample}.

\subsection{Graduated Reward Design}
\label{sec:reward}

\paragraph{The Sparse Reward Problem.}
In multi-step tool orchestration, standard binary rewards (Success/Failure) fail to provide adequate learning signals. This failure stems from two factors. First, the \textit{combinatorial explosion} of the action space: if a workflow requires $T$ steps and the probability of correctness at any step is $p < 1$, the probability of receiving a positive reward is $p^T$, which vanishes rapidly as $T$ increases. Second, the \textit{credit assignment ambiguity}: a binary reward of 0 does not distinguish between a model that failed at step 1 (wrong tool) and a model that succeeded for $T-1$ steps but failed at step $T$ due to a minor parameter typo. Without intermediate feedback, the agent faces a ``needle in a haystack'' optimization landscape where gradients are flat almost everywhere.

\paragraph{Two-Component Reward Structure.}
We decompose correctness into two complementary components that capture different aspects:
\begin{itemize}
    \item \textbf{$R_{\text{atomic}}$} evaluates each predicted action $y_t = (f_t, \boldsymbol{\theta}_t)$, i.e., whether the model generates syntactically and semantically valid function calls.
    \item \textbf{$R_{\text{orch}}$} evaluates workflow completion against $\mathbf{y}^*$, i.e., whether all required steps are executed in the correct order with dependencies respected.
\end{itemize}

\paragraph{$R_{\text{atomic}}$: Atomic Validity.}
For each predicted call, we assess correctness through two complementary validations:

\textit{Abstract Syntax Tree (AST) Validation.} We check syntactic correctness via three-level graduated scoring:
\begin{itemize}
    \item Level 1 ($\alpha_1$): Correct function name
    \item Level 2 ($\alpha_2$): Correct parameter structure and types
    \item Level 3 ($\alpha_3$): Exact parameter value matching
\end{itemize}
For a predicted call $\hat{y} = (\hat{f}, \hat{\boldsymbol{\theta}})$ matched against ground truth $y^* = (f^*, \boldsymbol{\theta}^*)$:
\begin{align}
R_{\text{AST}}(\hat{y}, y^*) &= \alpha_1 \cdot \ind[\hat{f} = f^*] + \alpha_2 \cdot s_{\text{struct}}(\hat{\boldsymbol{\theta}}, \boldsymbol{\theta}^*) \notag + \alpha_3 \cdot \ind[\hat{\boldsymbol{\theta}} = \boldsymbol{\theta}^*]
\end{align}
where $s_{\text{struct}}$ measures parameter overlap and type accuracy (see \Cref{app:reward-details} in the appendix).

\textit{Semantic Validation.} We execute the call through our deterministic environment and check for success: $R_{\text{sem}}(\hat{y}) = \ind[\text{execution succeeds}]$. This is execution-based validation rather than output matching: we reward successful execution regardless of whether the returned data exactly matches ground truth, catching errors that pass AST checks but fail at runtime (e.g., invalid date ranges, non-existent IDs).

\textit{Combined Score.} AST and semantic validations are complementary: AST rewards structural correctness while semantic rewards functional correctness. We combine them equally and average over all $K$ predicted calls:
\begin{equation}
R_{\text{atomic}} = \frac{1}{K} \sum_{k=1}^{K} \frac{R_{\text{AST}}(\hat{y}_k) + R_{\text{sem}}(\hat{y}_k)}{2}
\end{equation}

\paragraph{$R_{\text{orch}}$: Orchestration Consistency.}
While $R_{\text{atomic}}$ assesses local correctness, $R_{\text{orch}}$ enforces the global causal structure. Let $\mathcal{G} = (V, E)$ be the dependency graph of the ground truth workflow $\mathbf{y}^*$, where $V = \{y_1^*, \dots, y_T^*\}$ and an edge $(j, i) \in E$ indicates that $\boldsymbol{\theta}_i^*$ depends on observation $o_j$.

We define a mapping function $\mu: V \to \{1, \dots, K\} \cup \{\emptyset\}$ that aligns each ground truth action $y_i^*$ to the index of the first matching call in the predicted sequence $\hat{\mathbf{y}}$. A match requires the correct function name (parameter values are validated separately by $R_{\text{atomic}}$, enabling the two components to provide orthogonal learning signals).

The orchestration reward credits steps whose causal dependencies are satisfied:
\begin{equation}
R_{\text{orch}} = \frac{1}{|V|} \sum_{i \in V} \ind[i \text{ is matched}] \cdot \prod_{(j, i) \in E} \ind[\mu(j) < \mu(i)]
\end{equation}
Here, $\ind[\cdot]$ is the indicator function. The term $\ind[\mu(j) < \mu(i)]$ ensures that dependency $j$ is not only executed but executed \textit{before} the dependent step $i$. The multiplicative product acts as a strict gate: a step receives zero orchestration credit if any of its prerequisites are missing or out of order, providing a strong gradient signal against causal violations.

The total reward combines both components:
\begin{equation}
R_{\text{total}} = \lambda \cdot R_{\text{atomic}} + (1-\lambda) \cdot R_{\text{orch}}
\end{equation}
where $\lambda \in [0, 1]$ is a hyperparameter controlling the trade-off between atomic correctness and orchestration quality.

\section{Experiments}
\label{sec:experiments}

\subsection{Setup}

We evaluate on ComplexFuncBench~\citep{complexfuncbench}, a benchmark specifically designed to assess multi-step tool orchestration. The benchmark comprises 1,000 test samples spanning five API domains (hotels, flights, car rentals, attractions, and taxis), with an average of 3.26 steps per query. We apply GRPO~\citep{shao2024deepseekmath} for reinforcement learning using the verl framework~\citep{sheng2024hybridflow}. We conduct training experiments on two base models: Qwen3-8B~\citep{yang2025qwen3} and Qwen2.5-7B-Instruct~\citep{Yang2024Qwen25TR}.
\textbf{Zero-shot} refers to the original model without any task-specific training: the model receives a system prompt with tool descriptions and the user query, without in-context examples or fine-tuning.

\paragraph{Evaluation Metrics.}
Let $\mathcal{D}$ denote the evaluation set. For each sample $i \in \mathcal{D}$, the ground truth sequence $\mathbf{y}^{*(i)}$ is organized into $N^{(i)}$ turns, where each turn contains one or more function calls (possibly parallel).

\textbf{Turn Accuracy} measures workflow completion progress by counting consecutive turns completed from the beginning:
\begin{equation}
\text{Turn Acc} = \frac{\sum_{i \in \mathcal{D}} N_{\text{succ}}^{(i)}}{\sum_{i \in \mathcal{D}} N^{(i)}}
\end{equation}
where $N_{\text{succ}}^{(i)}$ is the number of consecutive successful turns for sample $i$. This metric reflects the cascading nature of multi-step orchestration: an early failure prevents all subsequent steps from succeeding. We also report \textbf{Call Accuracy} (proportion of correctly matched individual calls) as a secondary metric; detailed definitions are provided in \Cref{app:eval-metrics} in the appendix.

\paragraph{Training Configuration.} All experiments are conducted on 8$\times$H100 GPUs. We select batch size 16 and rollout size 16 based on preliminary experiments for stable training over extended steps. The Kullback-Leibler (KL) divergence coefficient is set to 0.001 (verl default), as we found it does not significantly affect performance. The reward weight $\lambda=0.5$ gives equal weighting to $R_{\text{atomic}}$ and $R_{\text{orch}}$.

\subsection{Effectiveness of RL Training Environment}
\label{sec:exp-oracle}

To isolate the effect of the RL environment from distribution shift, we first design an ``oracle'' experiment where models are trained and evaluated on the \textit{same} 100 queries from ComplexFuncBench. As shown in \Cref{tab:rl-effectiveness}, both models improve substantially, with gains of 14--17\% in turn accuracy. This validates that the cache-based environment and graduated reward provide effective learning signals. Note that baseline numbers differ from \Cref{tab:synthetic-data} due to the smaller evaluation set.

\begin{table*}[t]
\begin{minipage}[t]{0.48\textwidth}
\centering
\caption{Oracle: training and evaluating on the same queries (Turn Acc \%)}
\label{tab:rl-effectiveness}
\begin{small}
\resizebox{\textwidth}{!}{
\begin{tabular}{lccc}
\toprule
\textbf{Model} & \textbf{Zero-shot} & \textbf{+RL} & \textbf{Improv.} \\
\midrule
Qwen3-8B & 37.5 & \textbf{52.1} & +14.6 \\
Qwen2.5-7B-Instruct & 16.7 & \textbf{33.8} & +17.1 \\
\bottomrule
\end{tabular}
}
\end{small}
\end{minipage}%
\hfill
\begin{minipage}[t]{0.48\textwidth}
\centering
\caption{Reward component ablation (Turn Acc / Call Acc \%)}
\label{tab:reward-ablation}
\begin{small}
\begin{tabular}{lcc}
\toprule
\textbf{Config} & \textbf{Turn Acc} & \textbf{Call Acc} \\
\midrule
$R_{\text{atomic}}$ only & 32.2 & 15.3 \\
$R_{\text{orch}}$ only & 37.5 & 37.6 \\
\textbf{Combined} & \textbf{52.1} & \textbf{49.5} \\
\bottomrule
\end{tabular}
\end{small}
\end{minipage}
\end{table*}

\subsection{Generalization with Synthetic Data}
\label{sec:exp-synthesis}

We next evaluate whether models trained on \textit{disjoint} synthetic data can generalize to unseen queries. We generate training data using the constrained synthesis pipeline (\Cref{sec:environment}), varying samples per workflow template (3 or 10), and evaluate on the full 1000-sample ComplexFuncBench test set. We compare three strategies: SFT on ground-truth trajectories, GRPO with graduated rewards, and SFT followed by GRPO.

\begin{table*}[t]
\caption{Training with Constrained Synthetic Data (Turn Acc \%). Numbers in column headers indicate samples per workflow template (e.g., SFT-3 = SFT with 3 samples/template, GRPO-10 = GRPO with 10 samples/template). Evaluated on full 1000-sample test set.}
\label{tab:synthetic-data}
\centering
\begin{small}
\resizebox{\textwidth}{!}{
\begin{tabular}{lccccccc}
\toprule
\textbf{Model-LR} & \textbf{Zero-shot} & \textbf{SFT-3} & \textbf{SFT-10} & \textbf{GRPO-3} & \textbf{GRPO-10} & \textbf{SFT+GRPO-3} & \textbf{SFT+GRPO-10} \\
\midrule
Qwen3-8B-1e-6 & 34.4 & 32.6 & 31.7 & \textbf{36.4} & 30.2 & 31.1 & 29.9 \\
Qwen3-8B-5e-6 & -- & -- & -- & 28.7 & 29.6 & 22.9 & 28.4 \\
Qwen2.5-7B-1e-6 & 13.4 & 11.5 & 13.5 & 25.8 & 23.9 & \textbf{27.1} & 16.0 \\
Qwen2.5-7B-5e-6 & -- & -- & -- & 24.1 & 21.3 & 16.5 & 12.1 \\
\bottomrule
\end{tabular}
}
\end{small}
\end{table*}

\paragraph{Results.}
\Cref{tab:synthetic-data} reveals several findings. SFT alone yields negligible gains or even degradation: error analysis (\Cref{tab:error-analysis}) shows that SFT substantially increases function selection errors and worsens query parameter extraction, suggesting that imitation learning on limited templates does not generalize to novel query phrasings. Per-domain analysis (see \Cref{tab:domain-analysis} in the appendix) corroborates this, as SFT overfits to a subset of domains while degrading on others. GRPO with fewer samples per template sometimes outperforms more data, and Qwen2.5-7B achieves the largest gain with SFT+GRPO-3. These results suggest that for strong base models, direct RL training may be more effective than SFT when data is scarce.

\subsection{Cross-Benchmark Generalization}
\label{sec:cross-benchmark}

A key question is whether improvements on ComplexFuncBench transfer to other tool-use benchmarks with different API ecosystems. We evaluate our best model (Qwen3-8B, GRPO-3) on BFCL v4~\citep{bfcl}, a comprehensive function calling benchmark spanning agentic, multi-turn, live, and non-live categories. Crucially, the BFCL agentic tasks involve entirely different APIs than our Booking.com training environment, including web search and memory management tools, providing a genuine test of cross-domain transfer.

\Cref{tab:bfcl-agentic} presents detailed results on the BFCL v4 \textit{Agentic} category, which evaluates multi-step tasks requiring sequential API orchestration across two distinct tool ecosystems: web search and memory management. Our model consistently improves across both ecosystems and evaluation modes: Web Search accuracy improves from 3.5\% to 10.5\% in FC mode, while Memory accuracy improves from 17.4\% to 21.7\%. The gains are consistent across sub-categories, including Memory Recursive Summarization (30.3\% to 36.1\% in FC, 27.7\% to 40.0\% in Prompt). Overall BFCL accuracy also improves (+1.8pp FC, +0.8pp Prompt; full results in \Cref{app:bfcl-details}), while non-agentic categories remain stable, confirming that RL training specifically enhances orchestration capabilities without degrading single-step performance.

\begin{table}[t]
\caption{BFCL v4 Agentic category breakdown (\%). Web Search and Memory are entirely different API ecosystems from the Booking.com training domain. \textbf{+Ours} denotes GRPO-3.}
\label{tab:bfcl-agentic}
\centering
\begin{small}
\resizebox{\columnwidth}{!}{
\begin{tabular}{ll c ccc cccc}
\toprule
& & & \multicolumn{3}{c}{\textbf{Web Search}} & \multicolumn{4}{c}{\textbf{Memory}} \\
\cmidrule(lr){4-6} \cmidrule(lr){7-10}
\textbf{Mode} & \textbf{Model} & \textbf{Overall} & \textbf{Agg.} & \textbf{Base} & \textbf{No Snip.} & \textbf{Agg.} & \textbf{KV} & \textbf{Vector} & \textbf{Rec.\ Sum.} \\
\midrule
\multirow{2}{*}{FC} & Qwen3-8B & 10.5 & 3.5 & 5.0 & 2.0 & 17.4 & 8.4 & 13.6 & 30.3 \\
 & Qwen3-8B+Ours & 16.1 & 10.5 & 13.0 & 8.0 & 21.7 & 13.6 & 15.5 & 36.1 \\
\midrule
\multirow{2}{*}{Prompt} & Qwen3-8B & 10.1 & 7.0 & 5.0 & 9.0 & 13.1 & 4.5 & 7.1 & 27.7 \\
 & Qwen3-8B+Ours & 15.4 & 10.0 & 11.0 & 9.0 & 20.9 & 6.5 & 16.1 & 40.0 \\
\bottomrule
\end{tabular}
}
\end{small}
\end{table}

\subsection{Reward Component Analysis}
\label{sec:reward-analysis}

To understand the contribution of each reward component, we conduct ablation studies comparing $R_{\text{atomic}}$ only, $R_{\text{orch}}$ only, and the combined reward.

\begin{wraptable}{r}{0.52\textwidth}
\centering
\caption{Effect of Data Scale (Turn Acc \%). Scale denotes target samples per workflow; actual totals are 67, 179, 283, 383.}
\label{tab:scaling}
\begin{small}
\begin{tabular}{lcccc}
\toprule
\textbf{Model-LR} & \textbf{1} & \textbf{3} & \textbf{5} & \textbf{7} \\
\midrule
Qwen3-8B-1e-6 & 31.0 & 32.4 & 32.5 & 31.1 \\
Qwen3-8B-5e-6 & 29.4 & 26.4 & 30.3 & 27.3 \\
Qwen2.5-7B-1e-6 & 19.0 & 19.0 & 22.2 & 22.3 \\
Qwen2.5-7B-5e-6 & 22.4 & 27.4 & 22.8 & 19.3 \\
\bottomrule
\end{tabular}
\end{small}
\end{wraptable}

\begin{figure}[t]
\centering
\includegraphics[width=1\columnwidth]{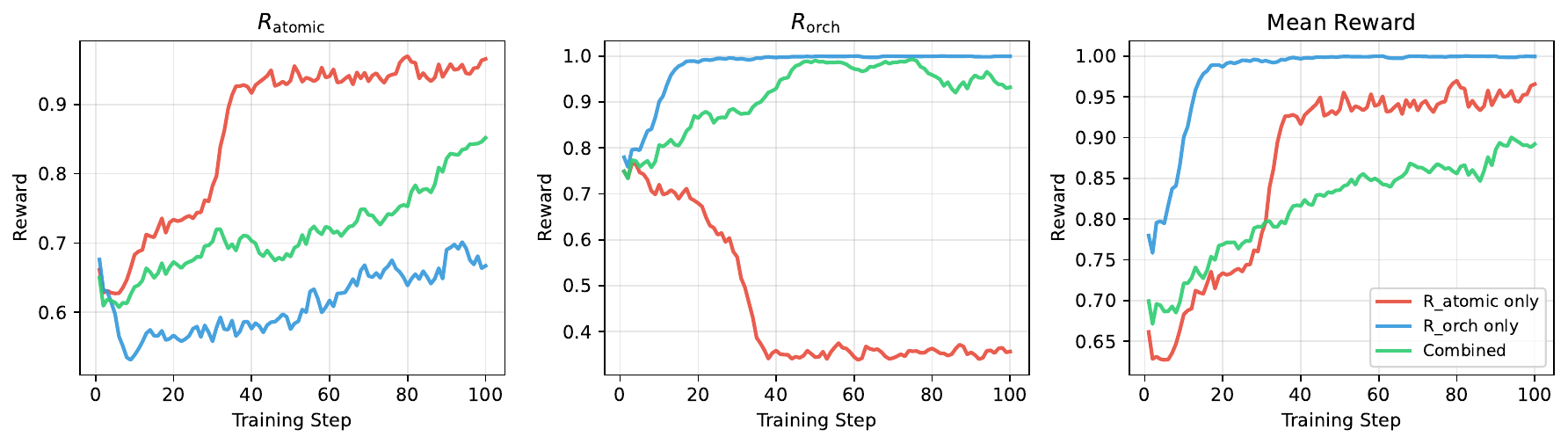}
\caption{Training dynamics under different reward configurations. \textbf{Left}: $R_{\text{atomic}}$ only achieves high atomic validity but orchestration collapses. \textbf{Middle}: $R_{\text{orch}}$ only achieves perfect orchestration but atomic validity drops. \textbf{Right}: Combined reward achieves balanced improvement.}
\label{fig:training-dynamics}
\vspace{-10pt}
\end{figure}

As shown in \Cref{tab:reward-ablation}, both components are essential: using either alone substantially degrades performance, while the combined reward achieves the best results. The training dynamics in \Cref{fig:training-dynamics} reveal why. Training with $R_{\text{atomic}}$ only leads to \textit{orchestration collapse}: the model achieves near-perfect individual call accuracy but loses the ability to coordinate multi-step sequences. Conversely, $R_{\text{orch}}$ only leads to \textit{atomic collapse}: the model completes workflows in the correct order but generates low-quality calls. Further analysis (see \Cref{app:training-analysis} in the appendix) confirms that under $R_{\text{orch}}$ only training, AST validity remains reasonable but semantic validity collapses, indicating the model generates syntactically correct but semantically broken calls.

\subsection{Scaling Analysis}
\label{sec:scaling}

We investigate data efficiency by varying the number of samples per workflow template. As shown in \Cref{tab:scaling}, performance plateaus quickly: a six-fold increase in data yields no consistent improvement, suggesting high data efficiency. The binding constraint appears to be template diversity rather than per-template sample density, as policy gradient methods learn from reward signals during exploration. Expanding template coverage may be more valuable than densifying existing templates.

\subsection{Stratified Performance Analysis}
\label{sec:stratified}

Overall accuracy improvements do not reveal \textit{where} RL training helps most. We stratify results by task complexity to understand which scenarios benefit from our approach.

\paragraph{Setting.} We analyze performance along two dimensions: (1) \textit{dependency depth}, the longest chain of sequential function calls where each depends on previous outputs, and (2) \textit{dependency pattern}, whether the workflow follows a linear chain or fans out into parallel branches. We compare three models: \textbf{Zero-shot} (no task-specific training), \textbf{GRPO-3} (trained on synthetic data, disjoint from test), and \textbf{RL-Oracle} (trained on test queries, representing the upper bound).

\begin{figure}[t]
\centering
\resizebox{0.8\textwidth}{!}{
\includegraphics[width=\columnwidth]{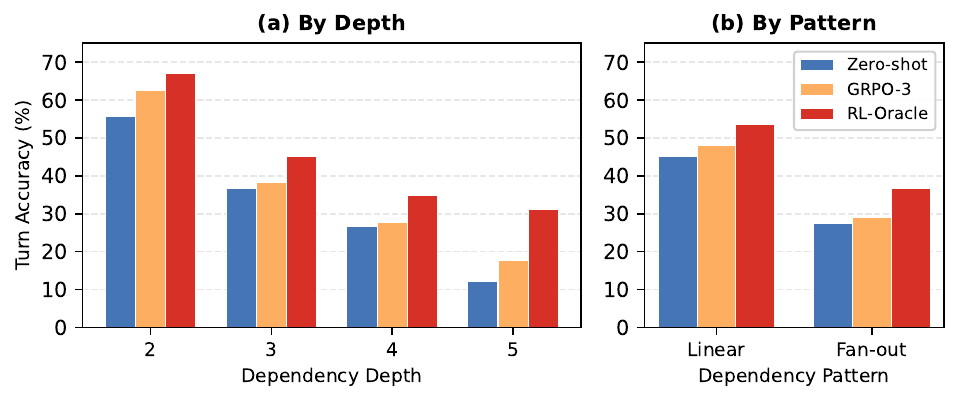}
}
\vspace{-4pt}
\caption{Turn Accuracy (\%) stratified by (a) dependency depth and (b) dependency pattern. Performance degrades with depth and for fan-out patterns. RL training provides consistent improvements across all stratifications.}
\label{fig:dep-analysis}
\vspace{-10pt}
\end{figure}

\Cref{fig:dep-analysis}(a) shows performance by dependency depth. Both GRPO-3 and RL-Oracle improve over Zero-shot, with relative gains increasing at greater depths (+20\% at depth 2, +31\% at depth 4 for RL-Oracle). \Cref{fig:dep-analysis}(b) shows that fan-out patterns are harder than linear chains, but RL training helps consistently. Additional stratification by logical composition is provided in \Cref{app:stratified-analysis} in the appendix.
\subsection{Error Analysis}
\label{sec:error-analysis}

To understand \textit{where} errors occur, we categorize failures at the call level (function name vs.\ parameter errors), parameter level (query-derived vs.\ dependency-propagated), and sequence level (workflow completion). Qualitative case studies are provided in \Cref{app:case-studies} in the appendix.

\Cref{tab:error-analysis} reveals several insights. Dependency propagation is not the primary bottleneck: dependency parameter errors are substantially lower than query parameter errors, suggesting the harder challenge is extracting parameters from natural language rather than propagating values between calls. The dominant failure mode is premature stopping after parameter errors; GRPO reduces all incomplete categories, particularly encouraging models to continue execution after correct steps. SFT substantially increases function selection errors while GRPO provides balanced improvements across all error types, suggesting imitation learning encourages more calls with lower accuracy whereas RL improves both quality and completion.

\begin{wraptable}{r}{0.52\textwidth}
\centering
\caption{Error analysis on ComplexFuncBench (Qwen3-8B). \textit{Call-level}: among produced calls; \textit{param-level}: given correct function; \textit{seq-level}: why incomplete samples stopped.}
\label{tab:error-analysis}
\begin{small}
\resizebox{0.5\columnwidth}{!}{
\begin{tabular}{lccc}
\toprule
\textbf{Error Type (\%)} & \textbf{Zero} & \textbf{SFT} & \textbf{GRPO} \\
\midrule
\multicolumn{4}{l}{\textit{Call-Level (among produced calls)}} \\
\quad Func.\ Selection Err & 15.0 & 15.3 & \textbf{14.8} \\
\quad Parameter Err & 18.7 & 18.2 & \textbf{16.5} \\
\midrule
\multicolumn{4}{l}{\textit{Parameter-Level (given correct func.)}} \\
\quad Query Param Err & 14.7 & 15.8 & \textbf{12.7} \\
\quad Dep.\ Param Err & 7.9 & \textbf{4.9} & 5.6 \\
\midrule
\multicolumn{4}{l}{\textit{Sequence-Level (incomplete breakdown)}} \\
\quad Stopped after Correct & 17.2 & 18.9 & \textbf{12.7} \\
\quad Stopped after Func Err & 12.9 & 28.3 & \textbf{16.6} \\
\quad Stopped after Param Err & 24.8 & 27.7 & \textbf{22.1} \\
\bottomrule
\end{tabular}
}
\end{small}
\end{wraptable}

\section{Related Work}
\label{sec:related}

\paragraph{Tool Learning Benchmarks and Environments.}
Comprehensive evaluations like BFCL~\citep{bfcl} and Gorilla~\citep{gorilla} have standardized tool-use assessment, though these often target multi-turn conversational scenarios with simple per-turn calls. Our work targets multi-step tool orchestration, where a single query necessitates complex, dependent API sequences. ComplexFuncBench~\citep{complexfuncbench} identifies parameter value errors as the dominant failure mode, motivating our graduated reward design; NESTFUL~\citep{nestful} and Seal-Tools~\citep{sealtools} further highlight performance degradation at depth $\geq 2$. Prior execution environments such as virtual API servers~\citep{stabletoolbench} and sandboxed setups~\citep{yao2024tau} primarily target stable evaluation with synthetic or domain-limited data. We extend this to the training phase by constructing a deterministic, cache-based environment with 100k+ real responses.

\paragraph{Training Data Generation.}
Generating high-quality multi-step traces remains a bottleneck. ToolAlpaca~\citep{toolalpaca} utilized multi-agent simulation, ToolLLM~\citep{toolllm} scaled to 16k APIs, and APIGen-MT~\citep{apigenmt} employs LLM committees for multi-stage verification. Our constrained synthesis pipeline achieves comparable success rates using a single LLM call by leveraging workflow templates as structural skeletons.

\paragraph{Reinforcement Learning for Tool Use.}
Standard RL approaches suffer from sparse binary rewards. ToolRL~\citep{toolrl} demonstrated PPO for single-step selection; ReTool~\citep{retool} applied iterative fine-tuning but focused on math/code rather than API orchestration; RAGEN~\citep{ragen} and PARL-MT~\citep{parlmt} identify the sparse reward issue but remain reliant on binary success signals. Our approach decomposes rewards into $R_{\text{atomic}}$ (structural validity) and $R_{\text{orch}}$ (dependency fulfillment) to provide dense, complementary learning signals.

\section{Conclusion}
\label{sec:conclusion}

We presented a framework for training LLMs on multi-step tool orchestration, validated in a controlled setting with Booking.com APIs: a deterministic cache-based environment with 100k+ real API responses, a workflow-aware data synthesis pipeline, and a graduated reward design combining $R_{\text{atomic}}$ and $R_{\text{orch}}$. Experiments on ComplexFuncBench demonstrate substantial improvements in turn accuracy, with ablations confirming both reward components are essential. Cross-benchmark evaluation on BFCL v4 shows that the learned orchestration skills transfer to entirely different API ecosystems (web search, memory management), with agentic task accuracy improving from 10.5\% to 16.1\%. Future work includes extending the framework to additional API domains and investigating automated template derivation from API specifications.

\section*{Ethics Statement}
This paper improves the reliability of LLMs in multi-step tool orchestration. As LLMs are integrated into autonomous agents executing real-world actions, ensuring they adhere to dependency constraints and parameter validity is crucial for safety. While our work reduces risks of incorrect API calls, autonomous agents carry inherent societal implications. Our constrained data synthesis and graduated rewards aim to provide more robust safeguards, contributing to safer agentic systems.


\bibliography{main}
\bibliographystyle{colm2026_conference}

\appendix

\section{Data Synthesis Algorithm}
\label{app:algorithm}

\begin{algorithm}[h]
   \caption{Constrained Data Synthesis Pipeline}
   \label{alg:synthesis}
\begin{small}
\begin{algorithmic}
   \STATE {\bfseries Input:} Workflow Templates $\mathcal{T}$, Cache $\mathcal{C}$, Generator LLM $\mathcal{M}$
   \STATE {\bfseries Output:} Synthetic Dataset $\mathcal{D}_\text{syn}$
   \STATE Build Inverted Index $\mathcal{I}: (f, \text{param}, \text{val}) \to \{\text{cache\_ids}\}$
   \FOR{each template $T = (f_1, \dots, f_n) \in \mathcal{T}$}
       \STATE Initialize empty trace $\tau = []$
       \FOR{step $t = 1$ {\bfseries to} $n$}
           \IF{step $t$ has dependencies on previous outputs}
               \STATE Extract required values $V$ from $\tau$
               \STATE Candidates $S_t \leftarrow \bigcap_{v \in V} \mathcal{I}(f_t, \textit{param}_v, v)$
           \ELSE
               \STATE Candidates $S_t \leftarrow \text{All entries for } f_t \text{ in } \mathcal{C}$
           \ENDIF
           \STATE Sample entry $(f_t, \boldsymbol{\theta}_t, o_t) \sim S_t$
           \STATE Append $(f_t, \boldsymbol{\theta}_t, o_t)$ to $\tau$
       \ENDFOR
       \STATE Extract $\mathbf{y}^* = \{(f_t, \boldsymbol{\theta}_t)\}_{t=1}^{n}$ from $\tau$
       \STATE Generate query $q \leftarrow \mathcal{M}(\text{``Generate user query...''}, \tau)$
       \STATE Validate $q$ by executing against environment
       \IF{success}
           \STATE Add $(q, \mathbf{y}^*)$ to $\mathcal{D}_\text{syn}$
       \ENDIF
   \ENDFOR
\end{algorithmic}
\end{small}
\end{algorithm}

\section{Training Sample Example}
\label{app:training-sample}

\begin{tcolorbox}[
      colback=gray!5, colframe=gray!60!black,
      coltitle=white, fonttitle=\small\bfseries,
      title=Example Training Sample (Car Rental Workflow),
      boxrule=0.8pt,
      left=2pt, right=2pt, top=2pt, bottom=2pt,
      arc=2pt
      ]
\textbf{Query $q$:} \textit{``I'm at the San Diego Marriott La Jolla. We're planning to rent a car for Halloween 2024, departing at 10 AM and returning at the same place at 10 AM the next day. Check the rental package price?''}

\textbf{Ground Truth $\mathbf{y}^* = (y_1^*, y_2^*, y_3^*)$:}
\begin{lstlisting}[language=json, basicstyle=\ttfamily\scriptsize]
Step 1: Search_Car_Location(query="San Diego Marriott La Jolla")
  -> o1: {coordinates: {lat: 32.87, lon: -117.22}}

Step 2: Search_Car_Rentals(pick_up_lat=32.87, pick_up_lon=-117.22,
          pick_up_date="2024-10-31", drop_off_date="2024-11-01", ...)
  -> o2: {search_results: [{vehicle_id: "637318066", ...}],
          search_context: {searchKey: <base64>}}

Step 3: Get_Packages(vehicle_id="637318066", search_key=<base64>)
  -> o3: {packages: [...]}
\end{lstlisting}
\textbf{Dependencies:} $y_2^*$ uses \texttt{lat, lon} from $o_1$; $y_3^*$ uses \texttt{vehicle\_id, search\_key} from $o_2$.
\end{tcolorbox}

\section{Additional Training Analysis}
\label{app:training-analysis}

This appendix provides additional training metrics that complement the main analysis in \Cref{sec:experiments}.

\begin{figure}[h]
\centering
\includegraphics[width=0.7\columnwidth]{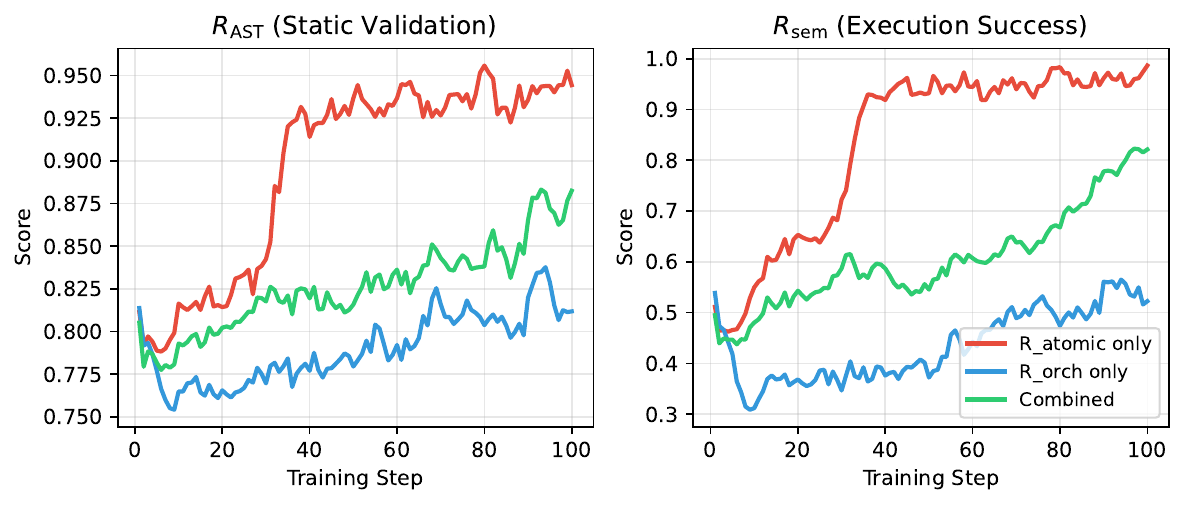}
\caption{Breakdown of $R_{\text{atomic}}$ into AST validation (static) and semantic validation (execution). $R_{\text{orch}}$ only training maintains $R_{\text{AST}}$ but $R_{\text{sem}}$ collapses, indicating syntactically valid but semantically broken calls.}
\label{fig:ast-vs-semantic}
\end{figure}

\begin{figure}[h]
\centering
\includegraphics[width=\columnwidth]{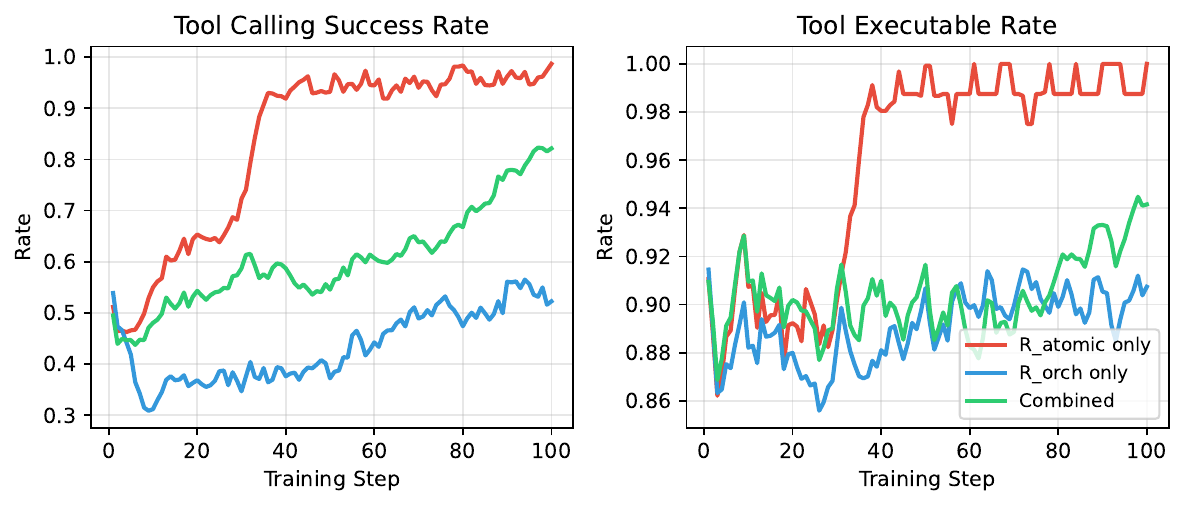}
\caption{Tool calling success metrics during training. \textbf{Left}: Tool calling success rate measures whether predicted calls match ground truth. \textbf{Right}: Tool executable rate measures whether calls execute without errors. $R_{\text{atomic}}$ only achieves highest success rates, while $R_{\text{orch}}$ only shows lower execution success despite high orchestration scores.}
\label{fig:success-rate}
\end{figure}

\begin{figure}[h]
\centering
\includegraphics[width=\columnwidth]{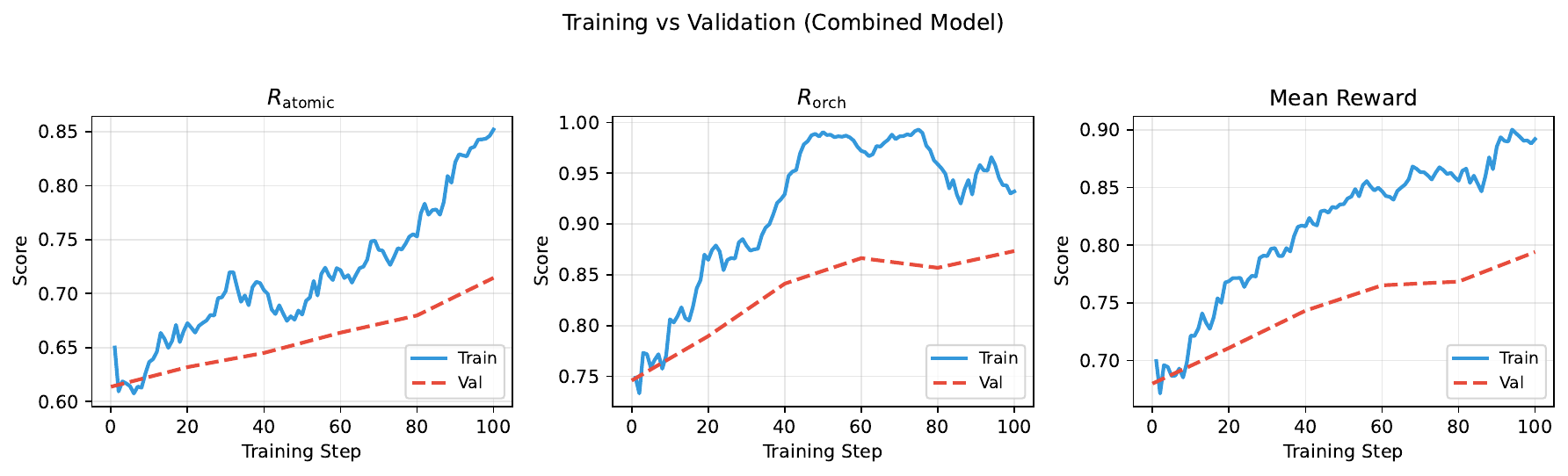}
\caption{Training vs validation metrics for the Combined model. All three metrics ($R_{\text{atomic}}$, $R_{\text{orch}}$, Mean Reward) show consistent improvement on both training and validation sets, with a moderate generalization gap indicating some overfitting to the training distribution.}
\label{fig:train-vs-val}
\end{figure}

\Cref{fig:success-rate} shows tool calling success metrics from a different perspective. The tool calling success rate (left) measures exact match with ground truth, while the executable rate (right) measures runtime success. Notably, $R_{\text{orch}}$ only training achieves only $\sim$52\% tool calling success despite $\sim$90\% executable rate, confirming that the model learns to generate \textit{different but executable} calls rather than the correct ones.

\Cref{fig:train-vs-val} demonstrates that the Combined model generalizes reasonably well, with validation metrics tracking training metrics throughout optimization. The gap between train and validation suggests opportunities for regularization or data augmentation in future work.

\section{Reward Computation Details}
\label{app:reward-details}

This section provides detailed definitions for the reward computation variables introduced in \Cref{sec:reward}.

\paragraph{Notation.} For a predicted call $\hat{y} = (\hat{f}, \hat{\boldsymbol{\theta}})$ and ground truth $y^* = (f^*, \boldsymbol{\theta}^*)$, we denote the parameter name sets as $\hat{\Theta} = \text{keys}(\hat{\boldsymbol{\theta}})$ and $\Theta^* = \text{keys}(\boldsymbol{\theta}^*)$, with $\hat{\Theta} \cap \Theta^*$ representing the parameters present in both calls.

\paragraph{Structure Score $s_{\text{struct}}$.} The structure score measures how well the predicted parameters match the expected schema:
\begin{equation}
s_{\text{struct}}(\hat{\boldsymbol{\theta}}, \boldsymbol{\theta}^*) = \underbrace{\frac{|\hat{\Theta} \cap \Theta^*|}{|\Theta^*|}}_{\text{coverage}} \cdot \underbrace{\frac{|\text{type-correct}|}{|\hat{\Theta} \cap \Theta^*|}}_{\text{type accuracy}}
\end{equation}
where:
\begin{itemize}
    \item \textbf{Coverage}: fraction of required parameters that are present in the prediction
    \item \textbf{Type accuracy}: among overlapping parameters, the fraction with correct types (e.g., string vs.\ integer, list vs.\ scalar)
    \item $|\text{type-correct}| = |\{p \in \hat{\Theta} \cap \Theta^* : \text{type}(\hat{\boldsymbol{\theta}}[p]) = \text{type}(\boldsymbol{\theta}^*[p])\}|$
\end{itemize}

\paragraph{Weight Values.} We use the following weights for graduated AST scoring:
\begin{itemize}
    \item $\alpha_1 = 1/3$: function name correctness
    \item $\alpha_2 = 1/3$: parameter structure and type correctness
    \item $\alpha_3 = 1/3$: exact parameter value matching
\end{itemize}
These weights sum to 1.0, ensuring $R_{\text{AST}} \in [0, 1]$.

\section{Workflow Template Example}
\label{app:workflow-template}

The following JSON structure illustrates a car rental workflow template with three steps and their dependency relationships:

\begin{tcolorbox}[
      colback=gray!5, colframe=gray!60!black,
      coltitle=white, fonttitle=\small\bfseries,
      title=Example Workflow Template (Car Rental),
      boxrule=0.8pt,
      left=2pt, right=2pt, top=2pt, bottom=2pt,
      arc=2pt
      ]
\begin{lstlisting}[language=json]
{"pattern": ["Search_Car_Location",
             "Search_Car_Rentals", "Get_Packages"],
 "dependencies": {
   "1": {"depends_on": [0],
         "dependency_args": {
           "pick_up_latitude": {"from_step": 0,
             "from_field": "[0].coordinates.latitude"},
           "pick_up_longitude": {"from_step": 0,
             "from_field": "[0].coordinates.longitude"}}},
   "2": {"depends_on": [1],
         "dependency_args": {
           "vehicle_id": {"from_step": 1,
             "from_field": "search_results[0].vehicle_id"},
           "search_key": {"from_step": 1,
             "from_field": "search_context.searchKey"}}}}}
\end{lstlisting}
\end{tcolorbox}

The template specifies: (1) an ordered function sequence, (2) a dependency graph where step 1 depends on step 0 (for coordinates) and step 2 depends on step 1 (for vehicle ID), and (3) field paths for extracting values from previous observations.

\section{Evaluation Metrics Details}
\label{app:eval-metrics}

\paragraph{Call Accuracy.} This secondary metric measures the proportion of individual function calls that are correctly matched, regardless of turn boundaries:
\begin{equation}
\text{Call Acc} = \frac{\sum_{i \in \mathcal{D}} K_{\text{correct}}^{(i)}}{\sum_{i \in \mathcal{D}} |\mathbf{y}^{*(i)}|}
\end{equation}
where $K_{\text{correct}}^{(i)}$ counts matched calls and $|\mathbf{y}^{*(i)}|$ is the total number of ground truth calls for sample $i$.

\paragraph{Matching Criteria.} A predicted call $\hat{y} = (\hat{f}, \hat{\boldsymbol{\theta}})$ matches ground truth $y^* = (f^*, \boldsymbol{\theta}^*)$ iff:
\begin{itemize}
    \item $\hat{f} = f^*$ (exact function name match)
    \item $\hat{\boldsymbol{\theta}} = \boldsymbol{\theta}^*$ (exact match on all parameter values)
\end{itemize}
This strict matching ensures that partial correctness (e.g., correct function but wrong parameters) is not credited.

\section{Stratified Analysis by Logical Composition}
\label{app:stratified-analysis}

\begin{table}[h]
\caption{Turn Accuracy (\%) stratified by logical composition type. The dataset is dominated by \textit{explicit\_conjunction} (86\%), with other patterns representing edge cases.}
\label{tab:logictype-analysis}
\centering
\begin{small}
\begin{tabular}{lcccc}
\toprule
\textbf{Type} & \textbf{Dist.} & \textbf{Zero-shot} & \textbf{GRPO-3} & \textbf{RL-Oracle} \\
\midrule
fallback\_logic & 5\% & 43.2 & 54.7 & 51.6 \\
explicit\_conj. & 86\% & 34.9 & 35.8 & 43.6 \\
alternative\_opt. & 4\% & 24.4 & 43.7 & 48.1 \\
parallel\_conj. & 5\% & 21.5 & 18.1 & 23.6 \\
\midrule
Overall & -- & 34.4 & 36.4 & 43.4 \\
\bottomrule
\end{tabular}
\end{small}
\end{table}

Among logical compositions, \textit{alternative\_options} (conditional branching) shows dramatic gains: GRPO-3 nearly doubles the Zero-shot baseline (43.7\% vs 24.4\%), with RL-Oracle reaching 48.1\%. Interestingly, for \textit{fallback\_logic}, GRPO-3 (54.7\%) slightly outperforms RL-Oracle (51.6\%), suggesting that synthetic data training may generalize better for certain compositional patterns.

\section{Per-Domain Performance Analysis}
\label{app:domain-analysis}

\begin{table}[h]
\caption{Turn Accuracy (\%) by API domain. Single-domain tasks (Attraction, Car-Rental) are generally easier than cross-domain tasks. SFT degrades cross-domain performance while GRPO maintains or improves across all domains.}
\label{tab:domain-analysis}
\centering
\begin{small}
\begin{tabular}{lccc}
\toprule
\textbf{Domain} & \textbf{Zero-shot} & \textbf{SFT-3} & \textbf{GRPO-3} \\
\midrule
Attraction & 49.8 & 57.9 & \textbf{60.7} \\
Car-Rental & 52.6 & 57.7 & \textbf{58.6} \\
Hotels & 25.4 & 22.6 & \textbf{29.0} \\
Flights & 33.4 & 29.4 & \textbf{33.6} \\
Cross-domain & \textbf{27.6} & 21.9 & 25.9 \\
\midrule
Overall & 34.4 & 32.6 & \textbf{36.4} \\
\bottomrule
\end{tabular}
\end{small}
\end{table}

\Cref{tab:domain-analysis} reveals domain-specific trends. Single-domain tasks (Attraction, Car-Rental) achieve 50--60\% turn accuracy, while cross-domain orchestration remains challenging at 22--28\%. Notably, SFT improves single-domain performance (Attraction: +8.1\%, Car-Rental: +5.1\%) but \textit{degrades} cross-domain performance (27.6\% to 21.9\%, $-$5.7\%), suggesting that SFT overfits to simpler patterns. In contrast, GRPO-3 maintains or improves performance across all domains.

\section{Prompt Templates}
\label{app:prompt-templates}

This section provides the prompt template used for query generation in our constrained data synthesis pipeline (\Cref{sec:method}).

\begin{tcolorbox}[
      colback=gray!5, colframe=gray!60!black,
      coltitle=white, fonttitle=\small\bfseries,
      title=Query Generation Prompt,
      boxrule=0.8pt,
      left=2pt, right=2pt, top=2pt, bottom=2pt,
      arc=2pt]
{\color{red!70!black}\textbf{System message:}}
\begin{lstlisting}[language=json, basicstyle=\ttfamily\scriptsize]
You are generating natural language queries for a travel booking assistant. You will be given EXACT API parameter values to use (from validated cache). Your task is ONLY to generate a natural language query that matches these exact parameters. DO NOT modify the parameter values.
\end{lstlisting}
\vspace{-0.5em}
\noindent\rule{\textwidth}{0.4pt}
\vspace{0.3em}
{\color{blue!70!black}\textbf{User message:}}
\begin{lstlisting}[language=json, basicstyle=\ttfamily\scriptsize]
Workflow pattern: {pattern}
EXACT PARAMETERS TO USE (do not modify):
Step 0 - {function_name}:
  {param}: '{value}' ...
Step 1 - {function_name}:
  (Parameters from previous step results) ...

Task: Generate a query matching these exact parameters.
OUTPUT FORMAT (JSON):
{"query": "...", "chosen_parameters": [...],
 "variation_notes": "Brief scenario description"}
IMPORTANT: Query must match the exact parameters above.
\end{lstlisting}
\end{tcolorbox}

The \texttt{chosen\_parameters} field serves as an echo-back mechanism for automatic consistency verification between the generated query and the underlying parameter bindings.

\vspace{1em}

\begin{tcolorbox}[
      colback=gray!5, colframe=gray!60!black,
      coltitle=white, fonttitle=\small\bfseries,
      title=Training/Inference System Prompt,
      boxrule=0.8pt,
      left=2pt, right=2pt, top=2pt, bottom=2pt,
      arc=2pt]
\begin{lstlisting}[language=json, basicstyle=\ttfamily\scriptsize]
You are a helpful assistant with access to booking.com APIs. You can help users search for hotels, flights, attractions, and more.

When you need to use a tool, output your tool call in this format:
<tool_call>
{"name": "function_name",
 "arguments": {"arg1": "value1", "arg2": "value2"}}
</tool_call>

After you make a tool call, you will receive a response:
<tool_response>
{"result": "..."}
</tool_response>

You may need to make multiple tool calls to complete a task. After gathering all necessary information, provide a helpful summary to the user.
\end{lstlisting}
\end{tcolorbox}

This system prompt is used during both RL training (rollouts) and SFT. Tool schemas are injected separately via the tokenizer's \texttt{apply\_chat\_template(tools=...)} parameter.

\section{Case Studies}
\label{app:case-studies}

This section presents case studies illustrating how RL training improves tool orchestration. Cases are selected where the baseline model fails but the RL-trained model succeeds.

\subsection{Alternative Options (Conditional Logic)}

The \textit{alternative\_options} category shows the largest improvement (24.4\% to 48.1\%). These tasks require handling conditional branches based on query results.

\begin{tcolorbox}[
      colback=gray!5, colframe=gray!60!black,
      coltitle=white, fonttitle=\small\bfseries,
      title=Case Study: Fallback Logic (Query Format Error),
      boxrule=0.8pt,
      left=2pt, right=2pt, top=2pt, bottom=2pt,
      arc=2pt
      ]
\textbf{Query:} \textit{``Check for anime activities in Tokyo with tickets after 12 PM on Nov 17. If not, check Osaka for tickets after 12 PM on Nov 20.''}

\textbf{Ground Truth:}
\begin{lstlisting}[language=json, basicstyle=\ttfamily\scriptsize]
1. Search_Attraction_Location({query: "Anime, Tokyo"})
2. Get_Availability({slug: "...", date: "2024-11-17"})
3. Search_Attraction_Location({query: "Anime, Osaka"})
\end{lstlisting}
\vspace{-0.5em}
\textbf{Zero-shot (0\% Turn Acc):} \texttt{Search\_Attraction\_Location(\{query: "Tokyo anime"\})} $\rightarrow$ \textcolor{red}{Query format mismatch}

\textbf{RL-Oracle (100\% Turn Acc):} Correctly executes all 3 steps with proper query format.
\end{tcolorbox}

The baseline model uses an incorrect query format (``Tokyo anime'' instead of ``Anime, Tokyo''), causing the search to fail. The RL-trained model learns the correct parameter format through execution feedback.

\subsection{Sequential Dependencies}

Sequential dependency tasks require correct parameter propagation between steps. The baseline often fails to use outputs from previous calls correctly.

\begin{tcolorbox}[
      colback=gray!5, colframe=gray!60!black,
      coltitle=white, fonttitle=\small\bfseries,
      title=Case Study: Sequential Dependencies (Parameter Propagation),
      boxrule=0.8pt,
      left=2pt, right=2pt, top=2pt, bottom=2pt,
      arc=2pt
      ]
\textbf{Query:} \textit{``Find car rentals at Paris CDG Airport and get supplier ratings for the top result.''}

\textbf{Ground Truth:}
\begin{lstlisting}[language=json, basicstyle=\ttfamily\scriptsize]
1. Search_Car_Rentals({pick_up_lat: 49.0, ...})
   -> {search_results: [...], search_context: {searchKey: "eyJ..."}}
2. Vehicle_Supplier_Ratings({vehicle_id: "774442521",
   search_key: "eyJkcml2ZXJz..."})
\end{lstlisting}
\vspace{-0.5em}
\textbf{Zero-shot (0\% Turn Acc):} Step 1 correct, but \textcolor{red}{failed to extract and propagate \texttt{search\_key}} to Step 2.

\textbf{RL-Oracle (100\% Turn Acc):} Correctly extracts \texttt{vehicle\_id} and \texttt{search\_key} from $o_1$ for Step 2.
\end{tcolorbox}

The RL-trained model correctly extracts and propagates the \texttt{search\_key} parameter, demonstrating improved inter-call dependency handling.

\subsection{Parallel Conjunction}

Parallel conjunction tasks require executing multiple independent operations and combining results.

\begin{tcolorbox}[
      colback=gray!5, colframe=gray!60!black,
      coltitle=white, fonttitle=\small\bfseries,
      title=Case Study: Parallel Conjunction (Multi-Domain Query),
      boxrule=0.8pt,
      left=2pt, right=2pt, top=2pt, bottom=2pt,
      arc=2pt
      ]
\textbf{Query:} \textit{``Recommend two top-rated activities and two highly-reviewed hotels in Montreal, Nov 20-23.''}

\textbf{Ground Truth:}
\begin{lstlisting}[language=json, basicstyle=\ttfamily\scriptsize]
1. Search_Attraction_Location({query: "Montreal"})
2. Search_Hotel_Destination({query: "Montreal"})
3. Search_Attractions({id: "...", sortBy: "score"})
4. Search_Hotels({dest_id: "...", ...})
\end{lstlisting}
\vspace{-0.5em}
\textbf{Zero-shot (0\% Turn Acc):} Only executes hotel branch; \textcolor{red}{missing parallel attraction search entirely}.

\textbf{RL-Oracle (100\% Turn Acc):} Executes both branches: hotels (Steps 1-2) and attractions (Steps 3-4).
\end{tcolorbox}

The baseline model completes only the hotel branch, missing the parallel attraction search. The RL-trained model learns to execute both branches of the conjunction.

\section{BFCL v4 Detailed Results}
\label{app:bfcl-details}

This section provides per-category breakdowns of the BFCL v4 results. The Agentic breakdown is presented in the main text (\Cref{tab:bfcl-agentic}).

\begin{table}[h]
\caption{BFCL v4 overall summary (\%). FC = Function Calling; Prompt = Prompt mode.}
\label{tab:bfcl-summary}
\centering
\begin{small}
\begin{tabular}{llcccccc}
\toprule
\textbf{Mode} & \textbf{Model} & \textbf{Overall} & \textbf{Agentic} & \textbf{Multi-Turn} & \textbf{Live} & \textbf{Non-Live} & \textbf{Halluc.} \\
\midrule
\multirow{2}{*}{FC} & Qwen3-8B & 41.6 & 10.5 & 41.6 & 80.4 & 88.8 & 80.5 \\
 & Qwen3-8B+Ours & 43.4 & 16.1 & 40.2 & 80.8 & 86.4 & 81.8 \\
\midrule
\multirow{2}{*}{Prompt} & Qwen3-8B & 38.7 & 10.1 & 31.6 & 80.1 & 89.3 & 82.4 \\
 & Qwen3-8B+Ours & 39.5 & 15.4 & 27.8 & 78.8 & 89.8 & 81.6 \\
\bottomrule
\end{tabular}
\end{small}
\end{table}

\begin{table}[h]
\caption{BFCL v4 Multi-Turn category breakdown (\%).}
\label{tab:bfcl-multiturn}
\centering
\begin{small}
\begin{tabular}{llccccc}
\toprule
\textbf{Mode} & \textbf{Model} & \textbf{Overall} & \textbf{Base} & \textbf{Miss Func} & \textbf{Miss Param} & \textbf{Long Ctx} \\
\midrule
\multirow{2}{*}{FC} & Qwen3-8B & 41.6 & 49.5 & 50.0 & 32.5 & 34.5 \\
 & Qwen3-8B+Ours & 40.2 & 48.5 & 44.0 & 36.5 & 32.0 \\
\midrule
\multirow{2}{*}{Prompt} & Qwen3-8B & 31.6 & 37.0 & 34.0 & 28.5 & 27.0 \\
 & Qwen3-8B+Ours & 27.8 & 32.5 & 33.0 & 25.0 & 20.5 \\
\bottomrule
\end{tabular}
\end{small}
\end{table}

\begin{table}[h]
\caption{BFCL v4 Non-Live AST category breakdown (\%).}
\label{tab:bfcl-nonlive}
\centering
\begin{small}
\begin{tabular}{llccccc}
\toprule
\textbf{Mode} & \textbf{Model} & \textbf{Overall} & \textbf{Simple} & \textbf{Multiple} & \textbf{Parallel} & \textbf{Par.\ Multi.} \\
\midrule
\multirow{2}{*}{FC} & Qwen3-8B & 88.8 & 75.2 & 96.0 & 94.0 & 90.0 \\
 & Qwen3-8B+Ours & 86.4 & 75.5 & 96.0 & 90.0 & 84.0 \\
\midrule
\multirow{2}{*}{Prompt} & Qwen3-8B & 89.3 & 77.8 & 96.5 & 93.5 & 89.5 \\
 & Qwen3-8B+Ours & 89.8 & 79.2 & 95.0 & 94.0 & 91.0 \\
\bottomrule
\end{tabular}
\end{small}
\end{table}

\begin{table}[h]
\caption{BFCL v4 Live AST category breakdown (\%).}
\label{tab:bfcl-live}
\centering
\begin{small}
\begin{tabular}{llccccc}
\toprule
\textbf{Mode} & \textbf{Model} & \textbf{Overall} & \textbf{Simple} & \textbf{Multiple} & \textbf{Parallel} & \textbf{Par.\ Multi.} \\
\midrule
\multirow{2}{*}{FC} & Qwen3-8B & 80.4 & 83.3 & 79.5 & 87.5 & 83.3 \\
 & Qwen3-8B+Ours & 80.8 & 82.6 & 80.4 & 81.2 & 75.0 \\
\midrule
\multirow{2}{*}{Prompt} & Qwen3-8B & 80.1 & 84.9 & 79.2 & 81.2 & 66.7 \\
 & Qwen3-8B+Ours & 78.8 & 83.0 & 78.2 & 75.0 & 66.7 \\
\bottomrule
\end{tabular}
\end{small}
\end{table}

\end{document}